\documentclass[11pt,a4paper]{article}
\usepackage[hyperref]{acl2020}
\usepackage{amsmath,graphicx}
\usepackage{times}
\usepackage{latexsym}
\usepackage{multirow}

\usepackage{url}

\usepackage{microtype}

\aclfinalcopy 

\title{Zero-Resource Cross-Domain Named Entity Recognition}

\author{Zihan Liu, Genta Indra Winata, Pascale Fung \\
Center for Artificial Intelligence Research (CAiRE)\\
Department of Electronic and Computer Engineering\\
The Hong Kong University of Science and Technology, Clear Water Bay, Hong Kong\\
\texttt{zihan.liu@connect.ust.hk}}

\date{}

\begin{document}
\maketitle
\begin{abstract}
Existing models for cross-domain named entity recognition (NER) rely on numerous unlabeled corpus or labeled NER training data in target domains. However, collecting data for low-resource target domains is not only expensive but also time-consuming. Hence, we propose a cross-domain NER model that does not use any external resources. We first introduce a \textit{Multi-Task Learning} (MTL) by adding a new objective function to detect whether tokens are named entities or not. We then introduce a framework called \textit{Mixture of Entity Experts} (MoEE) to improve the robustness for zero-resource domain adaptation. Finally, experimental results show that our model outperforms strong unsupervised cross-domain sequence labeling models, and the performance of our model is close to that of the state-of-the-art model which leverages extensive resources.
\end{abstract}

\section{Introduction}
Named entity recognition (NER) is a fundamental task in text understanding and information extraction. Recently, supervised learning approaches have shown their effectiveness in detecting named entities~\cite{ma2016end,chiu2016named,winata2019hierarchical}. However, there is a vast performance drop for low-resource target domains when massive training data are absent. To solve this data scarcity issue, a straightforward idea is to utilize the NER knowledge learned from high-resource domains and then adapt it to low-resource domains, which is called cross-domain NER. 

Due to the large variances in entity names across different domains, cross-domain NER has thus far been a challenging task. Most existing methods consider a supervised setting, leveraging labeled NER data for both the source and target domains~\cite{yang2017transfer, lin2018neural}.

However, labeled data in target domains is not always available.
Unsupervised domain adaptation naturally arises as a possible way to circumvent the usage of labeled NER data in target domains.
However, the only existing method, proposed by \citet{jia2019cross}, requires an external unlabeled data corpus in both the source and target domains to conduct the unsupervised cross-domain NER task, and such resources are difficult to obtain, especially for low-resource target domains. Therefore, we consider unsupervised zero-resource cross-domain adaptation for NER which only utilizes the NER training samples in a single source domain.

\begin{figure*}[!t]
    \centering
    \includegraphics[scale=0.83]{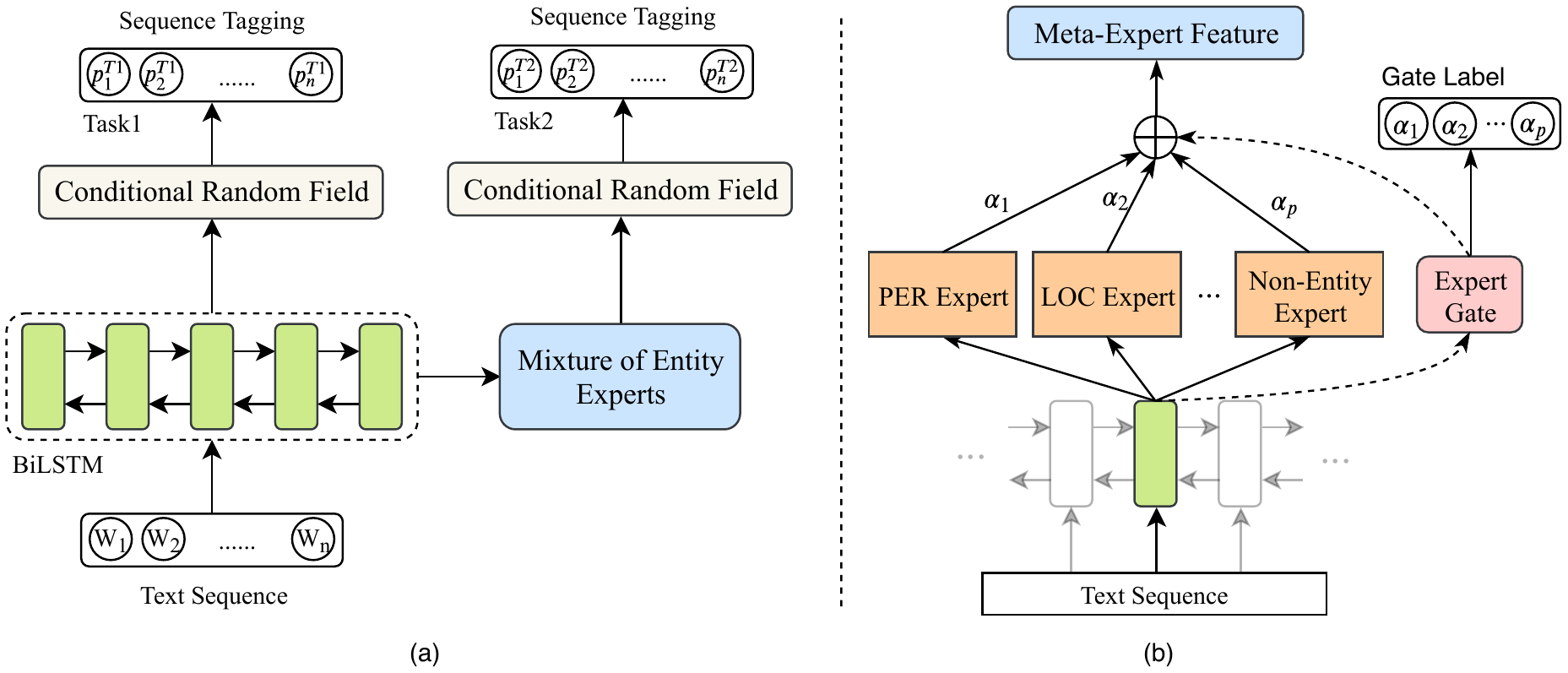}
    \caption{Model architecture (a) with multi-task learning and (b) with the Mixture of Entity Experts module.}
    \label{fig:model}
\end{figure*}

To meet the challenge of zero-resource cross-domain adaptation, we first propose to conduct multi-task learning (MTL) by adding an objective function to detect whether tokens are named entities or not. This objective function helps the model to learn general representations of named entities and to distinguish named entities from sequences in target domains.
In addition, we observe that in many cases, different entity categories could have a similar or the same context.
For example, in the sentence ``Arafat subsequently cancelled a meeting between Israeli and PLO officials,'' the person entity ``Arafat'', can be replaced with an organization entity within the same context, which illustrates the confusion among different entity categories and makes zero-resource adaptation much more difficult. 
Intuitively, when the entity type of a token is hard to be predicted based on the token itself and the token's context, we want to borrow the opinions (i.e., representations) from different experts. Hence, we propose a Mixture of Entity Experts (MoEE) framework to tackle the confusion of entity categories, and the predictions are based on the tokens and the context, as well as all entity experts. 

Experimental results show that our model is able to outperform current strong unsupervised cross-domain sequence tagging approaches, and reach comparable results to the state-of-the-art unsupervised method that utilizes extensive resources.

\section{Related Work}
Most of the existing work on cross-domain NER has been to investigate the supervised setting, where both source and target domains have labeled data~\cite{daume2007frustratingly,Obeidat2016LabelEA,yang2017transfer,lee2018transfer}. \citet{yang2017transfer} jointly trained models on the source and target domain with shared parameters. \citet{lin2018neural} added adaptation layers on top of existing models, and \citet{wang2018label} introduced label-aware feature representations for NER adaptation. \citet{lee2018transfer} utilized the idea of transfer learning by first initializing a target model with parameters learned from source-domain NER, and then using labeled target domain data to fine-tune the model. 
However, no prior work has focused on the unsupervised setting of cross-domain NER, except for~\citet{jia2019cross}. In \citet{jia2019cross}, however, external unlabeled data corpora resources in both the source and target domains are required to train language models for domain adaptations. This limitation has motivated us to develop a model that doesn't need any external resources.

Tackling the low-resource scenario where there are zero or minimal existing resources has always been an interesting yet challenging task~\cite{xie2018neural,liu2019attention,lample2017unsupervised,conneau2017word,shah2019robust}. Instead of utilizing large amounts of bilingual resources, \citet{liu2019zero,liu2019attention} only utilized a few word pairs for zero-shot cross-lingual dialogue systems. Unsupervised machine translation approaches~\cite{lample2017unsupervised,artetxe2017unsupervised} have also been introduced to circumvents the need of parallel data. \citet{winata2020learning} introduced the cross-accent speech recognition task and utilized meta-learning to cope with the data scarcity issue in target accents. 
\citet{bapna2017towards} and \citet{shah2019robust} proposed to do cross-domain slot filling with minimal resources.
To the best of our knowledge, we are the first to propose methods on cross-domain adaptation for NER with zero external resources.

\section{Methodology}
As illustrated in Fig.~\ref{fig:model}, our model combines a bi-directional LSTM and conditional random field (CRF) into a BiLSTM-CRF structure~\cite{lample2016neural} with MTL and MoEE modules. The parameters of BiLSTM are shared in the multi-task learning. 

\subsection{Multi-Task Learning}
Due to the large variations of named entities across domains, unsupervised cross-domain NER models often suffer from an inability to recognize named entities.
Hence, we propose to learn general representations of named entities and enhance the robustness for adaptation by adding an objective function to predict whether tokens are named entities or not, which is represented as $\textnormal{Task}_1$ in Fig.~\ref{fig:model}(a). To do so, based on the original named entity labels for each token in the training set, we create another label set, which represents whether tokens are named entities or not. Specifically, in this process, all non-entity tokens are consistent with the original labels, and other tokens belonging to different entity categories are classified as being in the same class representing the general named entity. $\textnormal{Task}_2$ in Fig.~\ref{fig:model}(a) represents the original NER task, which is to predict a concrete category for each token. Let us denote $X = [w_1, w_2, ..., w_n]$ as the input text sequence, and the MTL can be formulated as:
\begin{align*}
    [h_1, h_2, ..., h_n] &= \textnormal{BiLSTM}([w_1, w_2, ..., w_n]), \\
    [p_1^{T_{1}}, p_2^{T_{1}}, ..., p_n^{T_{1}}] &= \textnormal{CRF}_1([h_1, h_2, ..., h_n]),\\
    [m_1, m_2, ..., m_n] &= \textnormal{MoEE}([h_1, h_2, ..., h_n]), \\
    [p_1^{T_{2}}, p_2^{T_{2}}, ..., p_n^{T_{2}}] &= \textnormal{CRF}_2([m_1, m_2, ..., m_n]),
\end{align*}

where \textnormal{CRF}$_1$ and \textnormal{CRF}$_2$ denote the CRF layers for $\textnormal{Task}_1$ and $\textnormal{Task}_2$, respectively, 
and $[p_1^{T_{1}}, p_2^{T_{1}}, ..., p_n^{T_{1}}]$ and $[p_1^{T_{2}}, p_2^{T_{2}}, ..., p_n^{T_{2}}]$ represent the corresponding predictions.

\subsection{Mixture of Entity Experts}
Traditional NER models make predictions based on the features of the tokens and the context. Due to the confusion among different entity categories, NER models could easily overfit to the source domain entities and lose generalization ability to the target domain. Therefore, we introduce an MoEE framework, as depicted in Fig.~\ref{fig:model}(b).
It combines representations generated by experts to produce the final prediction. In this way, the knowledge from different experts is incorporated to model the inherent confusion and improve the generalization ability to target domains.


Each entity category acts as an entity expert, which consists of a linear layer. Note that we consider the non-entity as a special entity category.
The \textit{expert gate} consists of a linear layer followed by a softmax layer, which generates the confidence distribution over entity experts. We use the gold labels in $\textnormal{Task}_2$ to supervise the training of the expert gate.
Finally, the meta-expert feature incorporates features from all experts based on the confidential scores from the expert gate. We formulate the MoEE module as follows:
\begin{align}
    [\text{expt}_i^1, \cdots, \text{expt}_i^E] &= [\text{L}^1(h_i),\cdots,\text{L}^E(h_i)], \\
    [\alpha_1, \cdots, \alpha_E] &= \text{Softmax}(\text{Linear}(h_i)), \\
    m_i &= \sum_{a=1}^E \alpha_a * \text{expt}_i^a,
\end{align}
where $m_i$ is the meta-expert feature for the $i$-th hidden state of the BiLSTM, where \text{expt} is the feature generated from the expert, and \text{L} denotes the linear layer. We show that the MoEE has $E$ experts following the number of entity categories plus the non-entity category.
The expert features are computed based on the BiLSTM hidden states, and the predictions are conditioned on the expert features and the hidden states, which makes cross-domain adaptation more robust.

\subsection{Optimization} 
During training, we optimize for $\textnormal{Task}_1$, $\textnormal{Task}_2$ and the expert gate with cross-entropy losses $\mathcal{L}^{task1}$, $\mathcal{L}^{task2}$ and $\mathcal{L}^{gate}$, respectively, as we detail below:
\begin{align}
    \mathcal{L}^{task1} &= \sum_{j=1}^J \sum_{k=1}^{|Y_j|} -\log(p_{jk}^{T_{1}} \cdot (y_{jk}^{T_{1}})^{T}), \\ 
    \mathcal{L}^{task2} &= \sum_{j=1}^J \sum_{k=1}^{|Y_j|} -\log(p_{jk}^{T_{2}} \cdot (y_{jk}^{T_{2}})^{T}), \\
    \mathcal{L}^{gate} &= \sum_{j=1}^J \sum_{k=1}^{|Y_j|} -\log(p_{jk}^{gate} \cdot (y_{jk}^{gate})^{T}),
\end{align}
where J and $|Y_j|$ denote the number of training data and the length of the tokens for each training sample, respectively; $p_{jk}$ and $y_{jk}$ denote the predictions and labels for each token, respectively; and the superscripts of $p_{jk}$ and $y_{jk}$ represent the tasks. Hence, the final objective function is to minimize the sum of all the aforementioned loss functions.

\section{Experiments}
\subsection{Dataset}
We take the CoNLL-2003 English NER data~\cite{sang2003introduction} containing 15.0K/3.5K/3.7K samples for the training/validation/test sets as our source domain. We take the dataset containing 2K sentences from SciTech News provided by~\citet{jia2019cross} as our target domain. The datasets in the source and target domains contain the same four types of entities, namely, PER (person), LOC (location), ORG (organization), and MISC (miscellaneous).

\subsection{Experimental Setup}
\paragraph{Embeddings} 
We test our approaches on the FastText word embeddings~\cite{bojanowski2017enriching} and the pre-trained model BERT~\cite{devlin2019bert}. 
Entity names in the target domain are likely to be out-of-vocabulary (OOV) words because they don't usually exist in the source domain training set. FastText word embeddings are able to leverage the subword information and avoid the OOV problem, and BERT can solve this problem by using the BPE encoding. We try both freeze and unfreeze settings for FastText embeddings in the training. And for the BERT model, we add different modules (e.g., MoEE) on top to do fine-tuning.

\begin{table}[!t]
\centering
\resizebox{0.49\textwidth}{!}{
\begin{tabular}{lccc}
\hline
\multicolumn{1}{l|}{\multirow{2}{*}{\textbf{Model}}} & \multicolumn{1}{c|}{\textbf{BERT}} & \multicolumn{2}{c}{\textbf{FastText}} \\ \cline{3-4} 
\multicolumn{1}{l|}{} & \multicolumn{1}{c|}{\textbf{Fine-tune}} & \multicolumn{1}{c|}{\textbf{unfreeze}} & \textbf{freeze} \\ \hline
\multicolumn{4}{l}{\textit{Baseline}} \\ \hline
\multicolumn{1}{l|}{Concept Tagger} & \multicolumn{1}{c|}{67.14} & \multicolumn{1}{c|}{62.34} & 66.86 \\
\multicolumn{1}{l|}{Robust Sequence Tagger} & \multicolumn{1}{c|}{67.31} & \multicolumn{1}{c|}{63.66} & 68.12 \\ \hline
\multicolumn{4}{l}{\textit{Zero-Resource}} \\ \hline
\multicolumn{1}{l|}{BiLSTM-CRF} & \multicolumn{1}{c|}{67.55} & \multicolumn{1}{c|}{63.18} & 68.21 \\
\multicolumn{1}{l|}{w/ MTL} & \multicolumn{1}{c|}{\textbf{68.76}} & \multicolumn{1}{c|}{64.62} & 69.35 \\
\multicolumn{1}{l|}{w/ MoEE} & \multicolumn{1}{c|}{68.06} & \multicolumn{1}{c|}{\textbf{65.24}} & 69.27 \\
\multicolumn{1}{l|}{w/ MTL and MoEE} & \multicolumn{1}{c|}{68.59} & \multicolumn{1}{c|}{64.88} & \textbf{69.53} \\ \hline
\multicolumn{4}{l}{\textit{Using high-resource data in source and target domains}} \\ \hline
\multicolumn{1}{l|}{\citet{jia2019cross}} & \multicolumn{3}{c}{73.59} \\ \hline
\end{tabular}
}
\caption{F1-scores on the target domain. Models are implemented based on the corresponding embeddings.}
\label{tab:results}
\end{table}

\paragraph{Baselines}
Since we are the first to conduct zero-resource cross-domain NER, we compare our approach with strong unsupervised cross-domain sequence labeling models under minimal resources. \textbf{Concept Tagger} was proposed by~\citet{bapna2017towards} to utilize entity descriptions for unsupervised cross-domain utterance slot filling, and \textbf{Robust Sequence Tagger}~\cite{shah2019robust} was introduced to combined both entity descriptions and a few examples from each entity category for the same unsupervised task. In addition, we also compare our approach with the following baselines \textbf{BiLSTM-CRF}~\cite{lample2016neural}, \textbf{BiLSTM-CRF w/ MTL}, and \textbf{BiLSTM-CRF w/ MoEE}, as well as with the state-of-the-art model of the unsupervised cross-domain NER from~\citet{jia2019cross} which utilizes a large corpus in both the source and target domains.

\paragraph{Training Details}
For FastText embeddings~\footnote{Available in \url{https://fasttext.cc/docs/en/pretrained-vectors.html}} based models, we use a BiLSTM with a 200-dimensional hidden state and two layers. The linear layer size for each entity expert is 200. An Adam optimizer with a learning rate of 1e-3, a batch size of 32, and a dropout rate of 0.3 are used to train our model. We utilize the binary models provided in FastText to obtain the embeddings for OOV words.
For BERT-based models, given the strong textual understanding ability of the BERT model, we remove the BiLSTM from the text encoder, and only linear layer is utilized for sequence labeling (i.e., CRF layer is removed)~\cite{devlin2019bert}.
As for the evaluation, we use the standard IOB (in-out-begin) format to calculate the F1-score.

\begin{figure}[t!]
    \centering
    \includegraphics[scale=0.75]{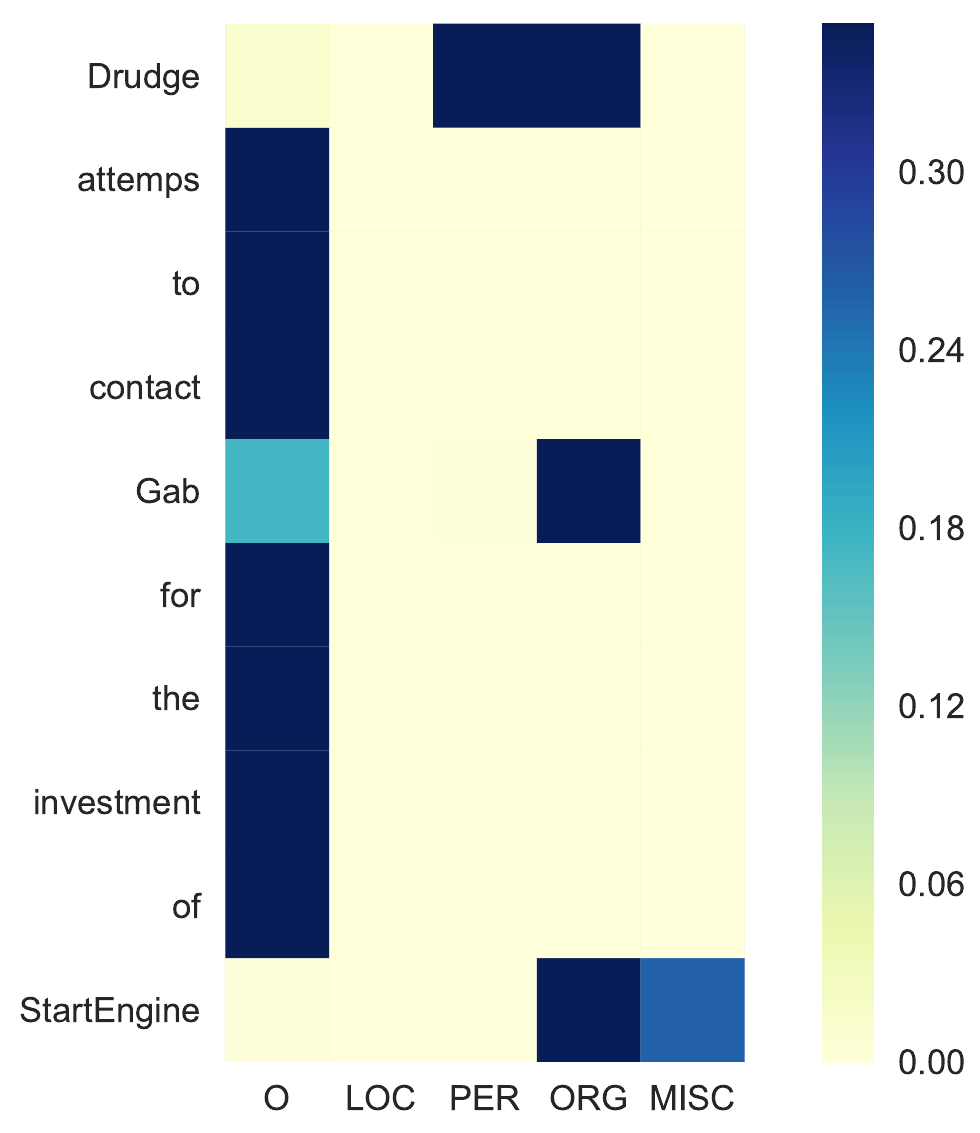}
    \caption{Confidence scores on different entity experts from the expert gate. ``O'' denotes non-entity expert.}
    \label{fig:visual}
\end{figure}

\subsection{Results \& Discussion}
From Table~\ref{tab:results}, our model combined with MTL and the MoEE outperforms the strong baselines Concept Tagger and Robust Sequence Tagger on all the embedding settings that we test. We conjecture that these two baselines, which utilize slot descriptions or slot examples, are suitable for limited slot names in the slot filling task, while they fail to cope with wide variances of entity names in the NER task across different domains, while our model is more robust to the domain variations. MTL helps our model recognize named entities in the target domain, while the MoEE adds information from different entity experts and helps our model detect the specific named entity types. Surprisingly, the performance of our best model (with freezed FastText embeddings) is close to that of the state-of-the-art model that needs a large data corpus in the source and target domains, which illustrates our model's generalization ability to the target domain.

We observe that the freezed FastText embeddings bring better performance than unfreezed ones. We conjecture that the embeddings could overfit to the source domain if we unfreeze them in the training.
Additionally, using freezed FastText embeddings is slightly better than BERT fine-tuning. We speculate that the reason is that NER is a word-level sequence tagging task, while the BERT model leverages subword embeddings, which could lose part of the word-level information for the task.

We visualize the confidence scores on different entity experts for each token in Fig.~\ref{fig:visual}. 
The expert gate can align non-entity tokens to the non-entity expert with strong confidence. For some entity tokens, e.g., ``Drudge'', the expert gate gives high confidence on more than one expert (e.g., ``PER'' and ``ORG'') since the model is not sure whether ``Drudge'' is a ``PER'' or ``ORG''. Our model is expected to learn the ``PER'' and ``ORG'' expert representations based on the hidden state of ``Drudge'', which contains the information of this token and its context, and then combine the expert representations for the prediction.

\section{Conclusion}
In this paper, we propose a zero-resource cross-domain framework for the named entity recognition task, which consists of multi-task learning and Mixture of Entity Experts modules.
The former learns the general representations of named entities to cope with the model's inability to recognize named entities, while the latter learns to combine the representations of different entity experts, which are based on the BiLSTM hidden states.
Experimental results show that our model outperforms strong cross-domain sequence tagging models, and the performance is close to that of the state-of-the-art model that utilizes extensive resources.

\section*{Acknowledgments}
This work is partially funded by ITF/319/16FP and MRP/055/18 of the Innovation Technology Commission, the Hong Kong SAR Government.

\bibliography{acl2020}
\bibliographystyle{acl_natbib}

\end{document}